\documentclass{article}

     \usepackage[nonatbib,final]{neurips_2020}

\usepackage[utf8]{inputenc} 
\usepackage[T1]{fontenc}    
\usepackage{hyperref}       
\usepackage{url}            
\usepackage{booktabs}       
\usepackage{amsfonts}       
\usepackage{nicefrac}       
\usepackage{microtype}      

\usepackage{amssymb}
\usepackage{mathtools}
\usepackage{booktabs}

\title{
\LARGE{Squirrel: A Switching Hyperparameter Optimizer}  \\\vspace{0.3cm} \Large{Description of the entry by AutoML.org \& IOHprofiler \\ to the NeurIPS 2020 BBO challenge}
}

\author{%
  Noor Awad $^1$%
  \And%
  Gresa Shala $^1$%
  \And%
  Difan Deng $^2$%
  \And%
  Neeratyoy Mallik $^1$%
  \And%
  Matthias Feurer $^1$%
  \And%
  Katharina Eggensperger $^1$%
  \And%
  Andr\'{e} Biedenkapp $^1$%
  \And%
  Diederick Vermetten $^3$%
  \And%
  Hao Wang $^3$%
  \And%
  Carola Doerr $^4$%
  \And%
  Marius Lindauer $^2$%
  \And%
  Frank Hutter $^{1,5}$%
  \AND%
  $^1$University of Freiburg $^2$Leibniz University Hannover $^3$Leiden University\\%
  $^4$Sorbonne Universit\'e, CNRS, LIP6 $^5$Bosch Center for Artificial Intelligence
}
\begin{document}
\maketitle
In this short note, we describe our submission to the NeurIPS 2020 BBO challenge. Motivated by the fact that different optimizers work well on different problems, our approach \emph{switches} between different optimizers.\footnote{Switching between algorithms also relates to work on \emph{chaining} or \emph{algorithm schedules}.} Since the team names on the competition's leaderboard were randomly generated  ``alliteration nicknames'', consisting of an adjective and an animal with the same initial letter, we called our approach the \emph{Switching Squirrel}, or here, short, \emph{Squirrel}. Our reference implementation of Squirrel is available at \url{https://github.com/automl/Squirrel-Optimizer-BBO-NeurIPS20-automlorg}.


The challenge mandated to suggest $16$ successive batches of $8$ hyperparameter configurations at a time. We chose to only use one optimizer for a given batch, warmstarted with all previous observations.\\
\textbf{In our Squirrel framework, we switched between the following components}: 
\begin{enumerate}
    \item An initial design (for known hyperparameter spaces: found by meta-learning; otherwise: selected by differential evolution) (3 batches);
    \item Optimization using Bayesian optimization by integrating the SMAC optimizer~\cite{hutter-lion11a,smac-2017} with a portfolio of different triplets of surrogate model, acquisition function, and output space transformation (8 batches); and
    \item Optimization using Differential Evolution (DE)~\cite{storn1997DE} with parameter adaptation (5 batches).
\end{enumerate}
\noindent{}We now discuss these components in turn.

%

 
\paragraph{Meta-learning the initial design}

Since the challenge was a blackbox challenge, it was a priori unknown which algorithms should be optimized using which configuration spaces. In order to mimic what one would do in a real-world optimization service based on the bayesmark~\cite{bayesmark} environment, for each of the configuration spaces provided by bayesmark we applied meta-learning to select an initial design of strong configurations to start from~\cite{reif-ml12a,feurer-aaai15a}. 
For each of these spaces, we performed extensive offline optimization on 11 datasets (the 6 datasets built into bayesmark and 5 additional OpenML~\cite{vanschoren-sigkdd13a} datasets\footnote{We only considered 5 additional datasets for lack of time and compute power; better performance could be obtained using more datasets for meta-learning.}), both for optimizing accuracy and negative log-likelihood. This resulted in 22 configurations\footnote{Extracted from the offline performance data using IOHanalyzer~\cite{iohanalyzer}.} for each of the configuration spaces in bayesmark.
When facing one of these known spaces (defined as an exact match of dimensions, names, and search ranges), we warm-started Squirrel with the corresponding 22 configurations, plus 2 randomly sampled ones, using 3 batches of 8 configurations.
%

When facing an unknown configuration space, Squirrel initialized the optimization procedure by employing DE with a random initial design, using DE for 3 batches of 8 configurations. 

\paragraph{Bayesian optimization with SMAC}

Squirrel exploits that the SMAC3 package implements different surrogate models (including GPs~\cite{rasmussen-book06a} \& random forests~\cite{breimann-mlj01a} (RF)) and different acquisition functions (Expected Improvement (EI)~\cite{jones-jgo98a}, LogEI~\cite{hutter-lion11a}, probability of improvement (PI)~\cite{Kushner_1964}, and lower confidence bound (LCB)~\cite{srninivas-icml10a}) that can be chosen from, making it easier to adapt to different loss landscapes and combat model mismatch. 
In view of the batch setting of the competition, Squirrel employs a portfolio of different triplets of surrogate model, acquisition function, and output space transformation to sample a batch of configurations in each iteration. This choice exploits that different acquisition functions implement different trade-off strategies between exploitation and exploration, while different surrogate models allow to fit different kinds of functions. Furthermore, Squirrel applies different transformations of the output space before fitting the surrogate model, incl. \texttt{log} and \texttt{copula}~\cite{pmlr-v119-salinas20a} transformations. 
By doing so, we warp the loss function landscape, which further promotes the diversity of the suggestions. 
For the portfolio, we made the following choices. GPs usually describe low-dimensional response functions better and EI is a very robust acquisition function; thus, we overall favoured GPs and EI. In total, we used $8$ different triplets as follows: GP + EI + no\_transform, GP + PI + no\_transform, GP + LCB + no\_transform, GP + EI + copula\_transform, GP + logEI + log\_transform, RF + EI + no\_transform, RF + logEI + logTransform and RF + EI + copula\_transform. These choices could likely be optimized further.    
To further promote diversity among the proposals we use the Kriging Believer~\cite{ginsbourger-cieop10a} and randomly reorder the triplets for each batch before sequentially querying them for suggestions.

\paragraph{Optimization with DE}
To complement the global search strategy of Bayesian optimization, Squirrel runs differential evolution (DE) as a more greedy and final stage for the last 5 batches.
Our version of DE takes the observations from the previous stages as an input to initialize a population of size 8 (the required batch size) by taking the best configuration observed so far, and choosing the other configurations of the population randomly from all previously-evaluated configurations. It then starts the evolutionary search by iterating three operations: mutation, crossover, and selection. For the mutation operation, we generated a child/offspring for each configuration in the population by the greedy $best/2$ mutation strategy \cite{storn1997DE}. 
The scaling factor $F$ which controls the difference vector in the mutation operation is adapted using a decreasing sinusoidal equation \cite{sin-de,Awad16}. 
\noindent{} This adaptation leads to more exploration of the search space when DE starts and more exploitation around the best-found region in later phases of the search. 
When the mutation phase is completed, the binomial (also known as uniform) crossover~\cite{storn1997DE} is applied to each target configuration $X_{i,g}$ and its corresponding mutant configuration $V_{i,g}$ to generate an offspring $U_{i,g}$. The crossover rate $CR_{i,g}$ is sampled from the normal distribution $\mathcal{N}(0.5,0.01)$, determining the probability by which a hyperparameter value is copied from the mutant configuration (the hyperparameter value is taken from the parent otherwise). 
After the final offspring is evaluated, truncation selection is applied: the trial configuration $U_{i,g}$ replaces its parent if it has a better function value and is discarded otherwise (in this case the parent configuration $X_{i,g}$ remains in the population).

\paragraph{Result in the official BBO competition}
Our submission ranked 3rd on the public leaderboard of the NeurIPS 2020 BBO challenge. For the final evaluation, it would have also ranked 3rd, had it not had a minor bug regarding the type of its output object, which let it crash.

In detail, the API specifications asked for the optimizers to return a list of dictionaries. Squirrel is built around a custom library for handling hyperparameter configurations~\cite{lindauer-arxiv19}, and we returned a list of its \texttt{Configuration} objects (a \texttt{Configuration} is a dict-like object implementing many methods of standard Python dictionaries, including all methods used by bayesmark). This worked with the challenge starter kit, the public bayesmark code and the testing server throughout the entire 3-months submission period without causing an error, but failed in the final evaluation, possibly because it does not fully implement the specifications of a Python dictionary. This indicates that the final evaluation used benchmarks with stricter type checks or different evaluation code than the code in the starter kit/bayesmark/test server.
After the official rankings, the organizers reran our code with a 1-line fix (which replaced our \texttt{Configuration} output object with its corresponding \texttt{dict} object); this fixed version yielded a score of 92.551495, which would have placed it on rank 3.

\paragraph{Result in the BBO competition's warmstarting-friendly leaderboard}
While the final evaluation used configuration spaces with altered hyperparameter names, the organizers also computed an alternate ``warmstarting-friendly leaderboard'' that uses un-altered hyperparameter names and thereby provides full information about the identity of the algorithm being optimized. Our submission won 1st place in this leaderboard (with a score of 94.845476 and the organizers' bootstrap analysis showing a 100\% confidence in this 1st place ranking), winning a prize of 3\,000 USD generously sponsored by 4Paradigm. This success underlines the power of even simple meta-learning approaches.


\bibliographystyle{plain}
\bibliography{strings,lib,local,proc}
\end{document}